\documentclass{article}
\usepackage[preprint,nonatbib]{neurips_2019}

\usepackage{amsthm, amsmath, amssymb}
\usepackage{mathtools}
\usepackage{amsfonts}
\usepackage{tikz}
\usetikzlibrary{positioning}
\usetikzlibrary{arrows}
\usepackage[edges]{forest}
\usepackage{microtype}
\usepackage{graphicx}
\usepackage{booktabs} 
\usepackage{mathabx}
\usepackage{pgfplots}
\pgfplotsset{compat=1.3}
\usepackage{pgfplotstable}
\pgfplotstableset{col sep=comma}
\usetikzlibrary{pgfplots.groupplots}
\usepackage{booktabs} 
\usepackage{filecontents}
\usepackage{multirow}
\usepackage{enumitem}
\usepackage{hyperref}
\usepackage{tikzscale}
\usepackage{physics}
\usepackage{adjustbox}

\pgfplotscreateplotcyclelist{exotic}{%
    orange,every mark/.append style={fill=orange!80!black},mark=square*\\%
    blue!60!white,every mark/.append style={fill=teal!80!black},mark=*\\%
    cyan!60!black,every mark/.append style={fill=cyan!80!black},mark=pentagon*\\%
    red!90!black,mark=otimes\\%
    lime!65!black,every mark/.append style={solid,fill=lime},mark=diamond*\\%
    red,densely dashed,every mark/.append style={solid,fill=red!80!black},mark=*\\%
    yellow!60!black,densely dashed,
    every mark/.append style={solid,fill=yellow!80!black},mark=square*\\%
    black,every mark/.append style={solid,fill=gray},mark=otimes*\\%
    blue,densely dashed,mark=star,every mark/.append style=solid\\%
    red,densely dashed,every mark/.append style={solid,fill=red!80!black},mark=diamond*\\%
}

\newtheorem*{rep@theorem}{\rep@title}
\newcommand{\newreptheorem}[2]{%
\newenvironment{rep#1}[1]{%
 \def\rep@title{#2 \ref{##1}}%
 \begin{rep@theorem}}%
 {\end{rep@theorem}}}
\makeatother

\newtheorem{lemma}{Lemma}
\newreptheorem{lemma}{Lemma}

\newtheorem{theorem}{Theorem}

\newtheorem{corollary}{Corollary}
\newreptheorem{corollary}{Corollary}
\newtheorem{definition}{Definition}

\newtheoremstyle{named}{}{}{\itshape}{}{\bfseries}{.}{.5em}{#1 \thmnote{\textbf{#3}}}
\theoremstyle{named}

\newcommand{\bdeu}[1]{\ensuremath{\mathrm{BDeu}(#1)}}
\newcommand{\pGam}[2]{\ensuremath{\Gamma_{#1}\left(#2\right)}}
\newcommand{\bdi}[2]{\ensuremath{\mathrm{LBDeu}(#1,#2)}}
\newcommand{\bd}[1]{\ensuremath{\mathrm{LBDeu}(#1)}}
\newcommand{\blbd}[2]{\ensuremath{\underline{h}(#1,#2)}}
\newcommand{\blbdh}{\ensuremath{\underline{h}}}
\newcommand{\blbdg}{\ensuremath{\underline{g}}}

\newcommand{\lbd}[2]{\ensuremath{\mathrm{LLBDeu}(#1,#2)}}

\newcommand{\pr}{\ensuremath{\mathrm{Pr}}}
\newcommand{\lbda}{\ensuremath{h}}
\newcommand{\D}{\ensuremath{\mathcal{D}}}
\newcommand{\ESS}{\ensuremath{\alpha_{\text{ess}}}}
\newcommand{\Vi}{\ensuremath{V^{\setminus i}}}
\newcommand{\U}{\ensuremath{\mathcal{D}_u}}
\newcommand{\LI}{\ensuremath{\mathcal{L}}}

\newcommand{\vect}[1]{\vec{#1}}

\newcommand{\getv}[2]{\ensuremath{#1^{#2}}}
\newcommand{\getd}[2]{\ensuremath{#1(#2)}}
\newcommand{\set}[1]{\ensuremath{\{{#1}\}}}

\forestset{
  colour my roots/.style={
    before typesetting nodes={
      edge=#1,
      for ancestors={
        edge=#1,
        #1,
      },
      #1,
    }
  },
  my edge label/.style={
    edge label={
      node [midway, fill=white, font=\large] {#1}
    }
  }
}

\title{On Pruning for Score-Based \\ Bayesian Network Structure Learning}

\author{%
  Alvaro H. C. Correia \footnotemark\\
  Eindhoven University of Technology \\
  Eindhoven, The Netherlands \\
  \texttt{a.h.chaim.correia@tue.nl}
  \And
  James Cussens \footnotemark[\value{footnote}]\\
  University of York\\
  York, United Kingdom \\
  \texttt{james.cussens@york.ac.uk}
  \And
  Cassio de Campos \footnotemark[\value{footnote}]\\
  Eindhoven University of Technology \\
  Eindhoven, The Netherlands\\
  \texttt{c.decampos@tue.nl}
}

\begin{document}
\maketitle
\begin{abstract}
    Many algorithms for score-based Bayesian network structure learning (BNSL), in particular exact ones, take as input a collection of potentially optimal parent sets for each variable in the data. Constructing such collections naively is computationally intensive since the number of parent sets grows exponentially with the number of variables. Thus, pruning techniques are not only desirable but essential. While good pruning rules exist for the Bayesian Information Criterion (BIC), current results for the Bayesian Dirichlet equivalent uniform (BDeu) score reduce the search space very modestly, hampering the use of the (often preferred) BDeu. We derive new non-trivial theoretical upper bounds for the BDeu score that considerably improve on the state-of-the-art. Since the new bounds are mathematically proven to be tighter than previous ones and at little extra computational cost, they are a promising addition to BNSL methods.
\end{abstract}

\label{sec:intro}
\section{Introduction}
A Bayesian network~\cite{pearl:88} is a widely used probabilistic graphical model. It is composed of (i) a \textit{structure} defined by a directed acyclic graph (DAG) where each node is associated with a random variable, and where arcs represent probabilistic dependencies entailing the {\em Markov} condition: every variable is conditionally independent of its non-descendant variables given its parents; and (ii) a collection of conditional probability distributions defined for each variable given its parents in the graph.  Their graphical nature make Bayesian networks ideal for complex probabilistic relationships existing in many real-world problems~\cite{doi:10.1002/gepi.21686}.

Bayesian network structure learning (BNSL) is essentially a search problem where we aim at finding the optimal graph given some data. We assume discrete data and tackle score-based learning, that is, we define the optimal graph as the structure maximising a data-dependent score~\cite{ml:heckermanetal}.
In particular, we focus on the \textit{Bayesian Dirichlet equivalent uniform} (BDeu) score~\cite{cooper92bayesian}, which consists in the log probability of the graph given (multinomial) data and a uniform prior on structures.
The BDeu score is decomposable; namely we can write it as a sum of \textit{local scores} of the domain variables:
$$
    \bdeu{\mathcal{G}} = \sum_{i\in V} \bdi{i}{S_i},
$$
where LBDeu is the local score function; $V=\{1,\ldots,n\}$ is the set of (indices of) variables in the
data, which are in correspondence with nodes of the network to be learned; and $S_i\subseteq \Vi$, with
$\Vi=V\setminus\{i\}$, the parent set of node $i$ in the DAG $\mathcal{G}$.

A common approach to \emph{exact} BNSL divides the problem into two steps: 
\begin{enumerate}
    \item \textsc{Candidate Parent Set Identification}: For each variable, find a suitable collection of candidate parent sets and their local scores.
    \item \textsc{Structure Optimisation}: Given the collection of candidate parent sets, choose a parent set for each variable so as to maximise the overall score while avoiding directed cycles.
\end{enumerate}
This paper concerns pruning ideas to help solve candidate parent set identification. Simply put, we aim at reducing the number of BDeu scores we have to compute by discarding parent sets that will not lead to an optimal solution at the second step.

BNSL is known to be NP-hard~\cite{chickering04:_large_sampl_learn_bayes_networ_np_hard} and the subproblem of parent set identification is unlikely to admit a polynomial-time (in $n$) algorithm; it is proven to be LOGSNP-Hard for BIC~\cite{koivisto:06}. As a compromise, one typically chooses a maximum in-degree $d$ (number of parents per node) and computes the score only for parent sets with in-degree at most $d$. 
Naturally, that does reduce the search space but comes at the cost of discarding numerous potentially optimal graphs. Conversely, increasing the maximum in-degree can considerably improve the chances of finding better structures but requires higher computing time: there are $\Theta(n^d)$ candidate parent sets (per variable) if an exhaustive search is performed with in-degree $d$, and $2^{n-1}$ without any in-degree constraint. The large search space is an important limiting factor in BNSL, as $d>2$ is already prohibitively expensive for many interesting applications~\cite{bartlett15:aij}. 

Our goal is then to prune this search space more aggressively to help scale exact BNSL with BDeu. We provide new theoretical upper bounds for the local scores that allow us to identify and discard non-optimal parent sets without ever having to compute their scores.
These new upper bounds are efficient and can be readily integrated to any searching approach~\cite{NIPS2016_6386,cussens:uai11,decampos2011a,campos09:_struc_learn_bayes_networ_const,jaakkola10:_learn_bayes_networ_struc_lp_relax,koivisto04:_exact_bayes_struc_discov_bayes_networ,Yuan12improved,yuan13:_learn_optim_bayes_networ}.
While our study has been motivated by the scientific interest in solving the BNSL problem in an exact manner, we shall note that local scores for a variable given its parents also have a probabilistic interpretation (the decomposition of the score comes from independence assumptions). Therefore, new approaches to prune such search space of parent sets can be useful for other purposes too.

The paper is organised as follows. Section~\ref{sect:def} provides the notation and required definitions, as well as a brief description of the current best bound for BDeu in the literature, which we call ub$_f$. Section~\ref{sect:naiveplus} presents a new improved bound ub$_g$ whose derivation follows the same mathematical approach as the existing state-of-the-art bound but further exploits properties of the score function to get better results. This new bound is provably tighter than previous ones but still does not capture all cases and other bounds can be devised. Section~\ref{sect:mlbased} looks at the problem from a new angle and introduces a bound ub$_h$ based on a (tweaked) maximum likelihood estimation. Bounds ub$_g$ and ub$_h$ leverage different aspects of the problem and we show how they can be effectively combined for an even more aggressive pruning in Section~\ref{sect:comb}. Finally, Section~\ref{sect:conc} concludes the paper and gives directions for future research. 

\section{Definitions and Notation}\label{sect:def}
First of all, since the collection of scores are computed independently for each variable in the dataset (BDeu is decomposable), we drop $i$ from the notation and use simply $\bd{S}$ to refer to the score of node $i$ with parent set $S\subseteq \Vi$.
We need some further notation:
\begin{itemize}[leftmargin=0.4cm, topsep=0pt]
  \setlength{\itemsep}{1pt}
  \setlength{\parskip}{1pt}
\item[$-$] The state space of variable $i$ is denoted by $c(i)$. Similarly, $c(S)$ is the set of all joint instantiations of the random variables in $S\subseteq V$, that is, $c(S)$ is the Cartesian product of the state space of involved variables,  $c(S)=\bigtimes_{i\in S} c(i)$. We denote the size of state space $c(S)$ as $q(S)=|c(S)|$, and we abuse notation to say $q(i)=|c(i)|$.

\item[$-$] We reserve $i$ for (indices of) variables and $j$ for instances of a state space, e.g., $j_{S} \in c(S)$. The subscript is omitted if clear from the context.
  
\item[$-$] The data $\D$ is a \textit{multiset} (repetitions are
  allowed) of elements from $c(V)$, with $\getv{\D}{S}$ the projection of $\D$ onto variables $S\subseteq V$ (note that $\D=\getv{\D}{V}$). The same notation applies for projections of instantiations, e.g. $\getv{j}{S}$. Moreover, we use $\getd{\getv{\D}{S}}{j_{S'}}\subseteq \getv{\D}{S}$ to denote the elements of $\getv{\D}{S}$ compatible with a given $j_{S'}\in c(S')$, that is, $\getd{\getv{\D}{S}}{j_{S'}}=\{j_{S} : j_{S} \in \getv{\D}{S}, \getv{j_S}{S\cap S'}=\getv{j_{S'}}{S\cap S'}\}$. Finally, we use $\U$ instead of $\D$ to denote the set of unique elements from a given multiset $\D$.

\item[$-$] For $j\in c(S)$, we define $n_{j}=|\getd{\getv{\D}{S}}{j}|$, that
  is, the number of occurrences of $j$ in $\getv{\D}{S}$. 

\item[$-$] The vector $\vect{\alpha}_j=(\alpha_{j,k})_{k\in c(i)}$ is the prior for parent set $S\subseteq \Vi$ under configuration $j\in c(S)$. In the BDeu score, $\vect{\alpha}_j$ satisfies $\alpha_{j,k}=\ESS/q(S\cup\{i\})$, where $\ESS$ is the equivalent sample size, a pre-defined user parameter to define the strength of
  the prior.
\end{itemize}

Let $\pGam{\alpha}{x}=\frac{\Gamma(x+\alpha)}{\Gamma(\alpha)}$ for $x$ non-negative integer and $\alpha>0$ ($\Gamma$ denotes the Gamma function). Denote $\sum_{k\in c(i)} \alpha_{j,k} = \ESS/q(S)$ by $\alpha_{j}$. The local score for $i$ with parent set $S\subseteq \Vi$ can be written as
\begin{flalign*}
    & \bd{S} =\!\! \sum_{j\in c(S)}\!\! \lbd{S}{j},\, \text{and}\\
    & \lbd{S}{j} \! = \! -\log \pGam{\alpha_{j}}{n_j} +\!\!\!\sum_{k\in c(i)}\! \log \pGam{\alpha_{j,k}}{n_{j,k}}\,.
\end{flalign*}
\noindent $\bd{S}$ is a sum of $q(S)$ values each of which is specific to a particular instantiation of variables in $S$. We call such values
\emph{local local BDeu scores (llB)}.
In particular, $\lbd{S}{j}=0$ if $n_j=0$, so we concentrate on instantiations $j$ that do appear in the data:
\begin{equation*}
    \bd{S} = \sum_{j\in \getv{\U}{S}} \lbd{S}{j}\, .
\end{equation*}
\noindent This formula does not come by chance. In Section~\ref{sect:mlbased} we discuss its relation with the posterior probability of having $S$ as parent of $i$.

\section{Pruning in Candidate Parent Set Identification}
\begin{figure}[t]
    \centering
        \adjustbox{max width=.9\textwidth, max height=\textheight}{
    \begin{forest}
    [\set{1}, for tree={grow=340, reversed=true, edge={thin}},
            [\set{1,2}, calign=last
                    [\set{1,2,3}, for tree={color=red}, for descendants={edge={red, densely dashed}}
                        [\set{1,2,3,4},
                            [\set{1,2,3,4,5}
                                [\set{1,2,3,4,5,6},
                                    [\set{1,2,3,4,5,6,7}]
                                ]
                                [\set{1,2,3,4,5,7}]
                            ]
                            [\set{1,2,3,4,6}
                                [\set{1,2,3,4,6,7}]
                            ]
                            [\set{1,2,3,4,7}]
                        ]
                        [\set{1,2,3,5}
                            [\set{1,2,3,5,6}
                                [\set{1,2,3,5,6,7}]
                            ]
                            [\set{1,2,3,5,7}]
                        ]
                        [\set{1,2,3,6}
                            [\set{1,2,3,6,7}]
                        ]
                        [\set{1,2,3,7}]
                    ]
                    [\set{1,2,4}
                        [\set{1,2,4,5}
                            [\set{1,2,4,5,6}
                                [\set{1,2,4,5,6,7}]
                            ]
                            [\set{1,2,4,5,7}]
                        ]
                        [\set{1,2,4,6}
                            [\set{1,2,4,6,7}]
                        ]
                        [\set{1,2,4,7}]
                    ] 
                    [\set{1,2,5}
                        [\set{1,2,5,6}
                            [\set{1,2,5,6,7}]
                        ]
                        [\set{1,2,5,7}]
                    ]
                    [\set{1,2,6}
                        [\set{1,2,6,7}]
                    ]
                    [\set{1,2,7}]
            ]
            [\set{1,3}, for tree={color=red}, for descendants={edge={red, densely dashed}}
                [\set{1,3,4}
                    [\set{1,3,4,5}, 
                        [\set{1,3,4,5,6}
                            [\set{1,3,4,5,6,7}]
                        ]
                        [\set{1,3,4,5,7}]
                    ]
                    [\set{1,3,4,6}
                        [\set{1,3,4,6,7}]
                    ]
                    [\set{1,3,4,7}]
                ]
                [\set{1,3,5}
                    [\set{1,3,5,6}
                        [\set{1,3,5,6,7}]
                    ]
                    [\set{1,3,5,7}]
                ]
                [\set{1,3,6}
                    [\set{1,3,6,7}]
                ]
                [\set{1,3,7}]
            ]
            [\set{1,4},
                [\set{1,4,5}
                    [\set{1,4,5,6}
                        [\set{1,4,5,6,7}]
                    ]
                    [\set{1,4,5,7}]
                ]
                [\set{1,4,6}
                    [\set{1,4,6,7}]
                ]
                [\set{1,4,7}]
            ]
            [\set{1,5}
                [\set{1,5,6}
                    [\set{1,5,6,7}]
                ]
                [\set{1,5,7}]
            ]
            [\set{1,6}
                [\set{1,6,7}]
            ]
            [\set{1,7}, l*=-1]
        ]
    \end{forest}
    }
    \caption{Illustration of potential parent sets in a dataset with 8 variables (the $8^{th}$ one is the child and does not show). This is still a small part of the search space with only sets including variable 1. In red dashed lines, the sets pruned if $\bd{\{1\}}\geq \text{ub}(\{1,3\})$.}
    \label{fig:csspn}
\end{figure}
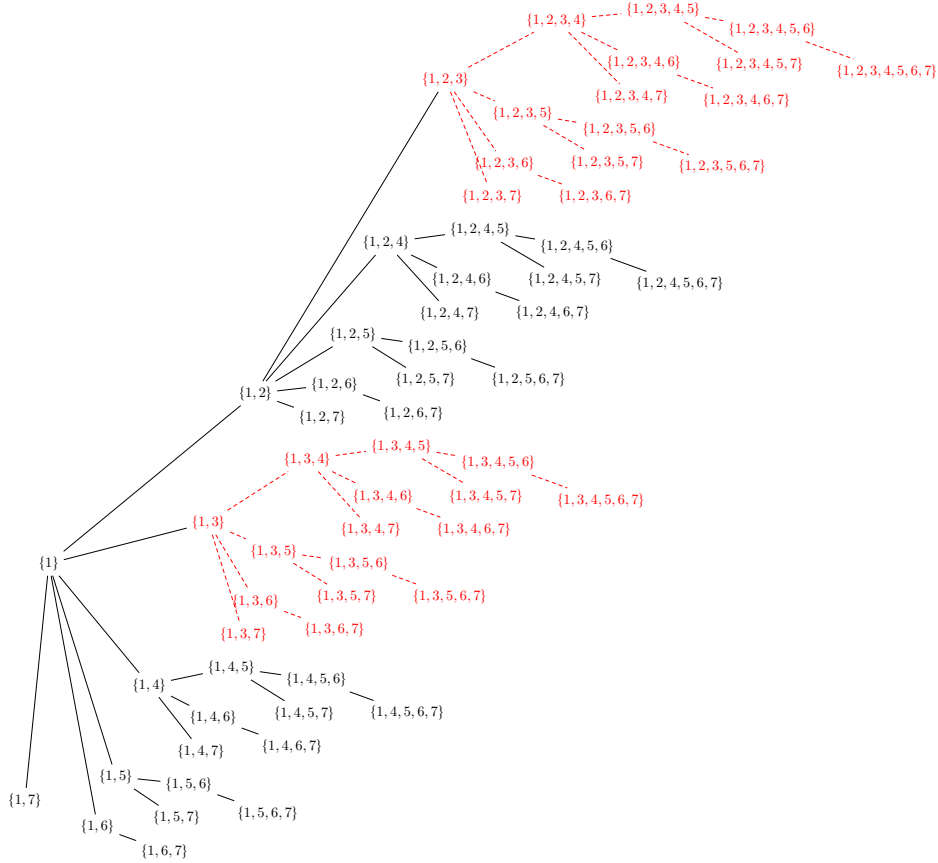

The pruning of parent sets rests on the (simple) observation that a parent set cannot be optimal if one of its subsets has a higher score \cite{teyssier05:_order}.
Thus, when learning Bayesian networks from data using BDeu, it is useful to have an upper bound 
\begin{equation}
    \mathrm{ub}(S) \geq \max_{T:T \supset S } \bd{T}\label{eq:tight}
\end{equation}
so as to potentially prune a whole area of the search space at once. Ideally, one would like an upper bound that is both tight (with respect to the inequality in Expression~\ref{eq:tight}) and cheap to compute, so that one can score parent sets incrementally, and at the same time check whether it is worth `expanding' them: if $\mathrm{ub}(S)$ is not greater than $\max_{R:R \subseteq S} \bd{R}$, then it is unnecessary to expand $S$. Figure~\ref{fig:csspn} illustrates how a hypothetical bound would prune the search space.

With that in mind, we can define candidate parent set identification more formally.
\begin{definition}[Candidate Parent Set Identification]\label{def:psi}
    For each variable $i \in V$,  find a collection of parent sets $$\LI_i = \{ S \subseteq \Vi : S'\subset S \Rightarrow \bd{S'} < \bd{S}\}\,.$$
\end{definition}
Unfortunately, we cannot predict the elements of $\LI_i$ and have to compute the scores for a list $L_i$ potentially much larger than $\LI_i$. The practical benefit of our bounds is to reduce $|L_i|$, thus lowering the computational cost of BNSL, while ensuring we do not miss any potentially optimal parent set, that is, $L_i\supseteq \LI_i$.
Before presenting the current best bound in the literature \cite{cussens12:_bdeu,campos10:_proper_bayes_diric_scores_learn,decampos2011a}, we give a lemma on the variation of counts with expansions of parent sets.

\begin{lemma}\label{lem:counts}
  For $S\subseteq T \subseteq \Vi$, $j_S\in\getv{\U}{S}$ and $j_T\in\getv{\U}{T}$ with $\getv{j_T}{S}=j_S$, we have $|\getv{\U}{T\cup\{i\}}| \geq |\getv{\U}{S\cup\{i\}}|,$ and  $|\getd{\getv{\U}{T\cup\{i\}}}{j_T}|\leq |\getd{\getv{\U}{S\cup\{i\}}}{j_S}|.$
  
\begin{proof}
Given that $S \subseteq T \subseteq \Vi$, every instantiation in $\getv{\U}{S\cup{i}}$ is compatible with one or more elements of $\getv{\U}{T\cup{i}}$, and thus $|\getv{\U}{T\cup{i}}| \geq |\getv{\U}{S\cup{i}}|$.
The relationship is reversed when we consider unique occurrences compatible with a given instantiation. By construction $\getv{j_T}{S}=j_S$, so if there is an instantiation $j_T\in\getv{\U}{T}$, there must be at least one corresponding $j_S\in\getv{\U}{S}$, and it follows that $|\getd{\getv{\U}{T\cup\{i\}}}{j_T}|\leq |\getd{\getv{\U}{S\cup\{i\}}}{j_S}|.$
Note that both $|\getd{\getv{\U}{T\cup\{i\}}}{j_T}|$ and $|\getd{\getv{\U}{S\cup\{i\}}}{j_S}|$ are bounded by $q(i)$: one instantiation for each value child $i$ can assume.
\end{proof}
\end{lemma}

As an example, consider the small dataset of Table \ref{tab:example}. The number of non-zero counts never decreases as we add a new variable to the parent set of variable $i=3$. With $S=\{1\}$ and $T=\{1,2\}$, we have $|\getv{\U}{S\cup\{i\}}| = 3$ and $|\getv{\U}{T\cup\{i\}}| = 4$. Conversely, the number of (unique) occurrences compatible with a given instantiation of the parent set never increases with its expansion: for example with $j_S=(1)$ and $j_T=(1,1)$, we have $|\getd{\getv{\U}{S\cup\{i\}}}{j_S}|=2$ and $|\getd{\getv{\U}{T\cup\{i\}}}{j_T}|=2$.

\begin{table}[ht]
\centering
    \caption{Example of data $\D$, its reductions by parent sets $S=\{1\}$ and $T=\{1,2\}$, and the number of unique occurrences compatible with $j_S\in\getv{\U}{S}$ and $j_T,j'_T\in\getv{\U}{T}$, with $\getv{j_T}{S}=\getv{{j'}_{T}}{S}=j_S$. The child variable is $i=3$, and we have $j_S=(1)$, $j_T=(1,1)$, $j'_{T}=(1,0)$.}
    \begin{adjustbox}{width=0.45\textwidth}
    \begin{tabular}{ccccc}
        \begin{tabular}[t]{ccc}
            \multicolumn{3}{c}{$\D$} \\
            \rule{0pt}{2ex}   
            $1$ & $2$ & $3$ \\
            \hline
            0 & 0 & 0 \\
            1 & 0 & 0 \\
            1 & 1 & 0 \\
            1 & 1 & 1 \\
        \end{tabular}
        \quad
        \begin{tabular}[t]{cc}
            \multicolumn{2}{c}{$\getv{\U}{S\cup\{i\}}$} \\
            \rule{0pt}{2ex}   
            $1$ & $3$ \\
            \hline
            0 & 0 \\
            1 & 0 \\
            1 & 1 \\
        \end{tabular}
        \quad
         \begin{tabular}[t]{ccc}
            \multicolumn{3}{c}{$\getv{\U}{T\cup\{i\}}$} \\
            \rule{0pt}{2ex}
            $1$ & $2$ & $3$ \\
            \hline
            0 & 0 & 0 \\
            1 & 0 & 0 \\
            1 & 1 & 0 \\
            1 & 1 & 1
        \end{tabular}
        \quad \\
        \begin{tabular}[t]{cc}
            \multicolumn{2}{c}{$\getd{\getv{\U}{S\cup\{i\}}}{j_S}$} \\
            \rule{0pt}{2ex}
            $1$ & $3$ \\
            \hline
            1 & 0 \\
            1 & 1
        \end{tabular}
        \quad
        \begin{tabular}[t]{ccc}
            \multicolumn{3}{c}{$\getd{\getv{\U}{T\cup\{i\}}}{j_T}$} \\
            \rule{0pt}{2ex}
            $1$ & $2$ & $3$ \\
            \hline
            1 & 1 & 0 \\
            1 & 1 & 1
        \end{tabular}
        \quad
        \begin{tabular}[t]{ccc}
            \multicolumn{3}{c}{$\getd{\getv{\U}{T\cup\{i\}}}{j'_{T}}$} \\
            \rule{0pt}{2ex}
            $1$ & $2$ & $3$ \\
            \hline
            1 & 0 & 0
        \end{tabular}
    \end{tabular}
    \end{adjustbox}
    \label{tab:example}
\end{table}

We now introduce function $f$ that, for a variable $i$, is defined on the sets of potential parents $S\subseteq\Vi$, and observed instantiations $j\in \getv{\U}{S}$:
\begin{equation}
    \begin{split}
        f(S,j) &= -|\getd{\getv{\U}{S\cup\{i\}}}{j}| \log q(i)\, , \\
        f(S) &= \sum_{j\in\getv{\U}{S}} f(S,j).
    \end{split}
\end{equation}
\begin{theorem}[ub$_f$]\label{thm:oldbound} For a variable $i$, a potential parent set $S\subseteq\Vi$ and its instantiations $j\in \getv{\U}{S}$, we have that $\lbd{S}{j} \leq f(S,j).$ 

Moreover, if $\bd{S'} \geq \sum_{j\in\getv{\U}{S}} f(S,j) = f(S)$ for some $S'\subset S$, then all $T\supseteq S$ are not in $\LI_i$ ~\cite{gobnilp:manual,decampos2011a}.
\end{theorem}

From Theorem~\ref{thm:oldbound}, we get an upper bound on the local BDeu score of all supersets of parent set $S$
\begin{equation}
    \text{ub}_f(S) = f(S) \geq \max_{T:T \supset S } \bd{T}.
\end{equation}

In words, we compute the number of non-zero counts per instantiation, $|\getd{\getv{\U}{S\cup\{i\}}}{j}|$, and we `gain' $\log q(i)$ for each of them. 
Note that $f(S) = -|\getv{\U}{S\cup\{i\}}|\log q(i)$, which by Lemma~\ref{lem:counts} is monotonically non-increasing over expansions of the parent set $S$. Hence $f(S)$ is not only an upper bound on $\bd{S}$ but also on $\bd{T}$ for every $T\supseteq S$.
Bound $\text{ub}_f$ is cheap to compute but is unfortunately too loose. We derive much tighter upper bounds on $\lbd{S}{j}$ (where $n_{j}>0$) by considering instantiation counts for the \textit{full} parent set $\Vi$, the parent set that includes all possible parents for child $i$. We call these \emph{full instantiation counts}. Evidently, the number of full parent instantiations $q(\Vi)$ grows exponentially with $|V|$, but it is linear in $|\D|$ when we consider only the unique elements $\getv{\U}{\Vi}$.

\section{Exploiting the Gamma Function}\label{sect:naiveplus}

First, we extend the current state-of-the-art upper bound of Theorem~\ref{thm:oldbound} by exploiting some properties of the Gamma function. For that, we need some intermediate results, where we assume $\alpha>0$.
\begin{lemma}\label{lem:little}
    Let $x$ be a positive integer. Then
    $$
    \pGam{\alpha}{0}=1 \text{ ~ and ~ } \log \pGam{\alpha}{x} = \sum_{\ell=0}^{x-1} \log (\ell + \alpha)\, .
    $$
    \begin{proof}
        Follows from $\Gamma(x+1) = x\Gamma(x)$.
    \end{proof}
\end{lemma}

\begin{lemma}
For $x$ positive integer and $v\geq 1$,
  \[
  \log(\frac{\pGam{\alpha}{x}}{\pGam{\alpha/v}{x}}) \geq \log v\, .
  \]
\label{lem:logv}
\end{lemma}\vspace{-0.8cm}
\begin{proof}
By applying Lemma~\ref{lem:little}, we obtain
\[
\sum_{\ell=0}^{x-1} \log\frac{\ell + \alpha}{\ell + \alpha/v} =
\log v+\sum_{\ell=1}^{x-1} \log\frac{\ell + \alpha}{\ell + \alpha/v} \geq \log v\, ,
\]
\noindent as each term of the sum (if any) is greater than zero.
\end{proof}
\begin{lemma}\label{lem:littlemed}
Let $x,y$ be non-negative integers such that $x+y>0$. Then
$$
  \left\{
  \begin{array}{@{}ll@{}}
    \pGam{\alpha}{x+y}=\pGam{\alpha}{x}\pGam{\alpha}{y} & \text{if}\ x\cdot y=0\, , \\
    \pGam{\alpha}{x+y} \geq \pGam{\alpha}{x}\pGam{\alpha}{y}\left(1 + y/\alpha \right) & \text{otherwise.}
  \end{array}\right.
$$
\begin{proof}
If $x$ (resp. $y$) is zero, then $\pGam{\alpha}{x}=1$ and the equality holds. Otherwise we apply Lemma~\ref{lem:little} three times and manipulate the products:
    \begin{equation}
        \begin{split}
            \frac{\pGam{\alpha}{x+y}}{\pGam{\alpha}{x}\pGam{\alpha}{y}} &=
            \frac{\prod_{z=0}^{x+y-1} (z+\alpha)}{\prod_{z=0}^{x-1} (z+\alpha)\prod_{z=0}^{y-1} (z+\alpha)} \nonumber \\
            &= \prod_{z=y}^{x+y-1}(z+\alpha)\prod_{z=0}^{x-1} \frac{1}{(z+\alpha)} \\
            &= \prod_{z=0}^{x-1}\frac{y+z+\alpha}{z+\alpha}\geq \frac{y+\alpha}{\alpha}\, ,
        \end{split}
    \end{equation}
    \noindent which holds since all terms in this final product are greater or equal to 1.
\end{proof}
\end{lemma}

\begin{corollary}\label{cor:l}
Let $x_1,\ldots,x_k$ be a list of non-negative integers in
  decreasing order with $x_1>0$, then
  \[
  \pGam{\alpha}{\sum_{l=1}^k x_l}\geq \prod_{l=1}^k \pGam{\alpha}{x_l} \prod_{l=1}^{k'-1} (1 + x_l/\alpha)\, ,
  \]
  \noindent where $k'\leq k$ is the last positive integer in the list (in this notation, the second product on the right-hand side disappears if $k'=1$).
  
\begin{proof}
  Repeatedly apply Lemma~\ref{lem:littlemed} to $x_t + (\sum_{l=t}^{k} x_l)$ until all elements are processed. While both the current $x_t$ and
  the rest of the list are positive (until $t=k'-1$), we obtain the extra term $(1 + x_t/\alpha)$. After that, we only `collect' the Gamma functions of the first product on the right-hand side, so the result follows.
\end{proof}
\end{corollary}

\begin{lemma}\label{theo:cassioloc}
For $S\subseteq \Vi$ and $j\in\getv{\U}{S}$, 
assume that $\vect{n}_{j}=(n_{j,k})_{k\in c(i)}$ are in decreasing order over $k=1,\ldots,q(i)$ (this is without loss of generality, since we can name and process them in any order). Then for any $\alpha\geq\alpha_j=\ESS/q(S)$, we have
\begin{equation*}
    \begin{split}
        \lbd{S}{j} \leq \text f(S,j)+g(S,j,\alpha), \\
        g(S,j,\alpha)=-\! \sum_{l=1}^{k'-1}\!\log\left(1 + n_{j,l}/\alpha\right),
    \end{split}
\end{equation*}
where $k'\leq k$ is the largest index such that $n_{j,k'}>0$.
\begin{proof}
First of all, we have
\begin{equation*}
\begin{split}
& \lbd{S}{j} = \! -\log \pGam{\alpha_{j}}{n_j} +\!\!\!\sum_{k\in c(i)}\! \log \pGam{\alpha_{j,k}}{n_{j,k}}\\
& = \! -\log \pGam{\alpha_{j}}{\sum_{k\in c(i)} n_{j,k}} +\!\!\!\sum_{k\in c(i)}\! \log \pGam{\alpha_{j,k}}{n_{j,k}}\, .
\end{split}
\end{equation*}
Since counts $n_{j,k}$ are in decreasing order by $k$, we apply Corollary~\ref{cor:l}:
\begin{equation*}
\begin{split}
    \lbd{S}{j}
    & \leq -\log\left(\prod_{l=1}^{q(i)} 
    \pGam{\alpha_j}{n_{j,l}} \prod_{l=1}^{k'-1} (1 + \frac{n_{j,l}}{\alpha_j})\right) + \sum_{k\in c(i)} \log\pGam{\alpha_{j,k}}{n_{j,k}} \\
    & =  \sum_{k\in c(i)} \log\left(\frac{\pGam{\alpha_{j,k}}{n_{j,k}}}{\pGam{\alpha_j}{n_{j,k}}}\right) - \sum_{l=1}^{k'-1}\log(1 + \frac{n_{j,l}}{\alpha_j}) \\
    & \leq -|\getd{\getv{\U}{S\cup\{i\}}}{j}| \log q(i) - \sum_{l=1}^{k'-1}\!\!\log(1\!+\! \frac{n_{j,l}}{\alpha})
\end{split}
\end{equation*}
\noindent with $\alpha\geq\alpha_j$ and 
$\pGam{\alpha_{j,k}}{n_{j,k}}/\pGam{\alpha_j}{n_{j,k}}\leq -\log q(i)$ by Lemma~\ref{lem:logv} whenever $n_{j,k}>0$.
\end{proof}

\end{lemma}
The difference here is the summation from the gap of the super-multiplicativity of $\Gamma$ (Lemma~\ref{lem:littlemed} and Corollary~\ref{cor:l}). That extra term gives us a tighter bound on \lbd{S}{j}, but $g(S) = f(S)+\sum_{j\in\getv{\U}{S}} g(S,j,\alpha)$ is no longer monotonic over expansions of $S$ (albeit monotone in $\alpha$). Hence, $g(S)$ is not an upper bound on $\bd{T}$ for every $T\supseteq S$, and we need further results on $g(S,j,\alpha)$. 
\begin{lemma}\label{theo:cassioprop}
For $S\subseteq T\subseteq \Vi$, $j_T\in\getv{\U}{T}$, and $j_S\in\getv{\U}{S}$ with $\getv{j_T}{S}=j_S$, we have $$f(T,j_T)\geq f(S,j_S),$$  $$g(T,j_T,\alpha) \geq g(S,j_S,\alpha).$$
\begin{proof}
Because $\getv{j_T}{S}=j_S$, $|\getd{\getv{\U}{T\cup\{i\}}}{j_T}|\leq |\getd{\getv{\U}{S\cup\{i\}}}{j_S}|$. Moreover, $n_{j_T,k}\leq n_{j_S,k}$ for every $k\in c(i)$ (the counts get partitioned as more parents are introduced to arrive at $T$ from $S$), so $(1 + n_{j_T,k}/\alpha) \leq (1 + n_{j_S,k}/\alpha)$ for every $k$, and the result follows.
\end{proof} 

\end{lemma}
Using this property of $g$ as described in Lemma~\ref{theo:cassioprop}, we can pick the best value of $g$ over all full expansions $j$ of a current instantiation $j_S$ to create a valid bound:

\begin{theorem}[ub$_g$]\label{thm:g}
Let $S\subseteq \Vi$, $j_S\in\getv{\U}{S}$, Then $$\lbd{S}{j_S} \leq f(S,j_S)+\blbdg(S,j_S)$$
$$\blbdg(S,j_S)=\min_{j\in\getv{\U}{\Vi}:~ \getv{j}{S}=j_S} g(\Vi,~j,~\ESS/q(S)).$$
Also, if $\bd{S'} \geq (f(S) + \sum_{j_S\in\getv{\U}{S}} \blbdg(S,j_S))=\blbdg(S)$ for some $S'\subset S$, then all $T\supseteq S$ are not in $\LI_i$.
\end{theorem}
\begin{proof}
First we prove that $f(S,j_S)+\blbdg(S,j_S)$ is an upper bound for $\lbd{S}{j_S}$.
From Lemma~\ref{theo:cassioprop}, if we take any instantiation of the fully expanded parent set, $j\in\getv{\U}{\Vi}:~ \getv{j}{S}=j_S$, we have that $g(S, j_S, \alpha) \leq g(\Vi, j, \alpha)$ for any $\alpha$.
As Lemma~\ref{theo:cassioprop} is valid for every full instantiation $j$, we take the minimum over them to get the tightest bound.
From Lemma~\ref{theo:cassioloc}, $\lbd{S}{j_S} \leq f(S,j_S)+\blbdg(S,j_S)$.
Now, if we sum all the llBs, we obtain the second part of the theorem for $S$.

Finally, we need to show that this second part of the theorem holds for any $T \supset S$, which follows from $f(T)\leq f(S)$ (as the total number of non-zero counts only increases, by Lemma~\ref{lem:counts}) and
\begin{equation*}
    \begin{split}
        \sum_{j_T\in\getv{\U}{T}} \blbdg(T,j_T) 
        & = \sum_{j_S\in\getv{\U}{S}} \left(\sum_{j_T\in\getv{\U}{T}:~ j_T^S=j_S} \blbdg(T,j_T)\right) leq \sum_{j_S\in\getv{\U}{S}} \blbdg(S,j_S)\, .
    \end{split}
\end{equation*}
\noindent That holds as $\blbdg(T,j_T)\leq 0$ and, with $j_T^S=j_S$, at least one term $\blbdg(T,j_T)$ is smaller than $\blbdg(S,j_S)$, as their minimisation spans the same full instantiations (and $g(\cdot,\cdot,\alpha)$ is non-decreasing on $\alpha$).
\end{proof}

In brief, the relevance of Theorem~\ref{thm:g} is that it gives us a tighter upper bound $\text{ub}_g(S) \leq \text{ub}_f(S)$, such that
\begin{equation*}
    \begin{split}
    \text{ub}_g(S) &= \blbdg(S) = (f(S) + \sum_{j_S\in\getv{\U}{S}} \blbdg(S,j_S)) \geq \max_{T:T \supset S } \bd{T}.
    \end{split}
\end{equation*}
Therefore, this bound is always equal or superior to the current state-of-the-art bound. Moreover,
the overhead of computing the bounds is negligible if a smart implementation is used (one that reuses computations that are nevertheless required for calculating the scores). The process which constructs contingency tables of counts for local score computations (say, from an AD-tree) is the main bottleneck in scoring, but it can be cheaply extended to simultaneously produce tables of sets of ``full instantiations'' for the computation of upper bounds where, for instance, addition of counts are replaced with unions of sets. While this technical detail is irrelevant for the mathematical proofs here, it is important to point out that the new bounds imply very little extra computational costs.

\section{Exploiting the Likelihood Function}
\label{sect:mlbased}

Bound $\text{ub}_g$ of previous section was based on the best full instantiation $j\in\getv{\U}{\Vi}$ that is compatible with an llB of the parent set $S$. Knowing that function $g$ is monotonic over parent set sizes, we could look at an instantiation of the fully extended parent set to derive a bound for the llB of $S$ and all its supersets. Even though the results are valid for every full instantiation, we can only compute bound $\text{ub}_g$ using one of them at a time.
The new bound of this section comes from the realisation that it is possible to exploit all full instantiations to derive a valid bound on the llB of $S$. For that purpose, we need some properties of inferences with the Dirichlet-multinomial distribution and conjugacy.

The BDeu score is simply the log marginal probability of the observed
data given suitably chosen Dirichlet priors over the parameters of a BN structure. Consequently, llBs are intimately connected to the Dirichlet-multinomial conjugacy. Given a Dirichlet prior $\vect{\alpha}_j=(\alpha_{j,1}, \dots, \alpha_{j,q(i)})$, the
probability of observing data $\D_{\vect{n}_j}$ with counts $\vect{n}_j=(n_{j,1}, \dots, n_{j,q(i)})$ is:
\begin{equation*}
    \label{eq:dmone}
    \log\pr(\D_{\vect{n}_j}|\vect{\alpha}_j) =
    \log\int_{p} \pr(  \D_{\vect{n}_j} |p) \pr(p|\vect{\alpha}_j) dp\, ,
\end{equation*}
where the first distribution under the integral is multinomial and the
second is Dirichlet. Note that
\begin{equation}
\log\int_{p} \pr( \D_{\vect{n}_j} |p) \pr(p|\vect{\alpha}_j) dp
\leq \max_{p} \log\pr(  \D_{\vect{n}_j} |p),
\label{eq:mlb}
\end{equation}
\noindent since $\int_{p} \pr(p|\vect{\alpha}_j) dp = 1$. Note also that llBs are not the probability of observing sufficient statistics counts, but of a particular dataset, that is, there is no multinomial coefficient which would consider all the permutations yielding the same sufficient statistics. Therefore, we may devise a new upper bound based on the maximum (log-)likelihood estimation.
\begin{lemma}\label{l:noname}
  Let $S\subseteq\Vi$ and $j\in \getv{\U}{S}$. Then
  $
  \lbd{S}{j} \leq \text{ML}(\vect{n}_{j})\, ,
  $
  where we have that 
  $\text{ML}(\vect{n}_{j})=\sum_{k\in c(i)} n_{j,k} \log (n_{j,k}/n_{j})$. (In this notation, we use $0\log 0 = 0$.)
  
\begin{proof}
The llB is simply the log probability of observing a data sequence with counts
$\vect{n}_j$ under a Dirichlet-multinomial distribution with parameter vector
$\vect{\alpha}_j$. The result follows from Expression~\eqref{eq:mlb}
and holds for any prior $\vect{\alpha}_{j}$.
\end{proof}

\end{lemma}

\begin{corollary}\label{cor:ll}
  Let $S\subseteq\Vi$ and $j_S\in \getv{\U}{S}$. Then
  $
  \lbd{S}{j_S} \leq \sum_{j\in \getv{\U}{\Vi}:~ j^S=j_S}\text{ML}(\vect{n}_{j})\, .
  $
\begin{proof}
  This follows from the properties of the maximum likelihood estimation, because it is monotonically non-decreasing with the expansion of parent sets (in terms of maximum likelihood, we fit the distribution just as well or better when having more parents).
\end{proof}
\end{corollary}

We can improve further on this bound of Corollary~\ref{cor:ll} by considering llBs as a function $\lbda$ of $\alpha$ for fixed $\vect{n}_j$, since we can study and exploit the shape of their curves. We define
\begin{equation*}
  \label{eq:fa}
  \lbda_{\vect{n}_j}(\alpha) = -\log \pGam{\alpha}{n_{j}}
  +     \sum_{k\in c(i)} \log
  \pGam{\alpha/q(i)}{n_{j,k}}\, .
\end{equation*}

\begin{lemma}
  \label{conj:todo}
  If $\not\exists k : n_{j,k} = n_{j}$, then $\lbda_{\vect{n}_j}$ is a concave function for positive $\alpha\leq 1$.
  
  \begin{proof}
  (This result can also be obtained from~\cite{levin77}.)
  Using the identity in Lemma~\ref{lem:little}, or, equivalently, by
  exploiting known properties of the digamma and trigamma functions, we have
  \begin{equation*}
    \label{eq:digamma}
    \pdv{\lbda_{\vect{n}_j}}{\alpha} (\alpha) = - \sum_{\ell=0}^{n_{j}-1} \frac{1}{\ell + \alpha} + \sum_{k=1}^{q(i)}\sum_{\ell=0}^{n_{j,k}-1} \frac{1}{\ell q(i) + \alpha}, \text{ and }
  \end{equation*}
  \begin{equation*}
    \label{eq:trigamma}
        \pdv[2]{\lbda_{\vect{n}_j}}{\alpha} (\alpha) = \sum_{\ell=0}^{n_{j}-1} \frac{1}{(\ell + \alpha)^{2}} - 
    \sum_{k=1}^{q(i)}\sum_{\ell=0}^{n_{j,k}-1} \frac{1}{(\ell q(i) + \alpha)^{2}}\, .
  \end{equation*}
  It suffices to show that $\pdv[2]{\lbda_{\vect{n}_j}}{\alpha} (\alpha)$ is always negative under the conditions of the theorem.
If there are at least two $n_{j,k}>0$, then
\begin{equation*}
  \pdv[2]{\lbda_{\vect{n}_j}}{\alpha} (\alpha) \leq \sum_{\ell=0}^{n_{j}-1} \frac{1}{(\ell + \alpha)^{2}} - \frac{2}{\alpha^2}
\end{equation*}
\noindent simply by ignoring all those negative terms with $\ell\geq 1$.
Now we approximate it by the infinite sum of quadratic reciprocals:
\begin{equation*}
    \begin{split}
         \pdv[2]{\lbda_{\vect{n}_j}}{\alpha} (\alpha) & \leq \sum_{\ell=0}^{n_{j}-1} \frac{1}{(\ell + \alpha)^{2}} - \frac{2}{\alpha^2} \\ & =  -\frac{1}{\alpha^2} + \frac{1}{(1+\alpha)^2} +\sum_{\ell=2}^{n_{j}-1} \frac{1}{(\ell + \alpha)^{2}} \\ 
         &< -\frac{1}{\alpha^2} + \frac{1}{(1+\alpha)^2} +\sum_{\ell=2}^{\infty} \frac{1}{\ell^{2}} \\
        & = -\frac{1}{\alpha^2} + \frac{1}{(1+\alpha)^2} + \frac{\pi^2}{6} - 1\, , \nonumber
    \end{split}
\end{equation*}
which is negative for any $\alpha\leq 1$ (the gap between the two fractions containing $\alpha$ obviously decreases with the increase of $\alpha$, so it is enough to check the sign for the largest value $\alpha=1$). Thus we have $\pdv[2]{\lbda_{\vect{n}_j}}{\alpha} (\alpha)<0$.
\end{proof}
\end{lemma}

The concavity of $\lbda_{\vect{n}_j}$ is useful for the following reason.
\begin{lemma}
  \label{thm:useconcave}
  Let $S\subseteq\Vi$ and $j\in \getv{\U}{\Vi}$ such that 
  $\not\exists k : n_{j,k} = n_{j}$. If $\alpha\leq q(S)$ and $\pdv{\lbda_{\vect{n}_j}}{\alpha} (\alpha/q(S))$ is
  non-negative, then $$\lbda_{\vect{n}_j}(\alpha/q(T)) \leq \lbda_{\vect{n}_j}(\alpha/q(S)) \text{ for every } T\supseteq S.$$
  
\begin{proof}
  Since $\not\exists k : n_{j,k} = n_{j}$ and $\alpha/q(S)\leq 1$, we have that $\lbda_{\vect{n}_j}$ is concave (Lemma~\ref{conj:todo}) and
  since $\pdv{\lbda_{\vect{n}_j}}{\alpha} (\alpha/q(S)) \geq 0$, $\lbda_{\vect{n}_j}$ is non-decreasing.
\end{proof}

\end{lemma}

The final step to improve the upper bound is to consider any local score of a parent set $S$ as a function of the (log-)probabilities over full mass functions.
\begin{lemma}
Let $S\subseteq\Vi$ and $j_S\in\getv{\U}{S}$. Then
$$
\lbd{S}{j_S}\leq \smashoperator{\sum_{j\in\getv{\U}{\Vi}:~ j\neq j^\star}} \text{ML}(\vect{n}_{j}) + \log\pr(\D_{\vect{n}_{j^\star}}|\vect{\alpha}_{j_S}),
$$
where
$j^\star=\arg\min_{j\in\getv{\U}{\Vi}}\log \pr(\D_{\vect{n}_{j}}|\vect{\alpha}_{j_S})$.  \label{thm:mlplus}
\end{lemma}
\begin{proof}
We rewrite $n_{j_S,k}$ as the sum of counts from full mass functions:
$
n_{j_S,k} = \sum_{j\in\getv{\U}{\Vi}:~\getv{j}{S}=j_S} n_{j,k}.
$
Thus, $\lbd{S}{j_S}$ is the log probability $\log\pr(\D_{\vect{n}_{j_S}}|\vect{\alpha}_{j_S})$ of observing a data sequence with counts $\vect{n}_{j_S}=(\sum_{j\in\getv{\U}{\Vi}:~\getv{j}{S}=j_S} n_{j,k})_{k\in c(i)}$ under the Dirichlet-multinomial with parameter vector $\vect{\alpha}_{j_S}$.
Assume an arbitrary order for the full mass functions related to elements in $\{j\in\getv{\U}{\Vi}:~\getv{j}{S}=j_S\}$ and name them $j_1,\ldots,j_{w}$, with $w=|\{j\in\getv{\U}{\Vi}:~\getv{j}{S}=j_S\}|$.
Exploiting the conjugacy multinomial-Dirichlet we can express this probability as a product of conditional probabilities:
\begin{equation*}
\pr(\D_{\vect{n}_{j_S}}|\vect{\alpha}_{j_S})=\prod_{\ell=1}^{w} \pr\left(\D_{\vect{n}_{j_{\ell}}}\Bigg|\sum_{t=1}^{\ell-1}\vect{n}_{j_t}+\vect{\alpha}_{j_S}\right)\, , 
\end{equation*}
\begin{equation*}
    \begin{split}
        \lbd{S}{j_S}
        & = \sum_{\ell=1}^{w}\log \pr\left(\D_{\vect{n}_{j_{\ell}}}\Bigg|\sum_{t=1}^{\ell-1}\vect{n}_{j_t}+\vect{\alpha}_{j_S}\right) \\
        & \leq \log \pr(\vect{n}_{j_1}|\vect{\alpha}_{j_S}) + \sum_{t=2}^w \text{ML}(\vect{n}_{j_t}).
    \end{split}
\end{equation*}
\noindent These are obtained by applying Expression~\eqref{eq:mlb} to all but the first term. Since the order is arbitrary, we can pick one in our best interest and the result follows.
\end{proof}

While the bound of Lemma~\ref{thm:mlplus} is valid for $S$, it gives no assurances about its supersets $T$, so it is of little direct use (if we need to compute it for every $T\supset S$, then it is better to compute the scores themselves). To address that, we replace the first term of the right-hand side summation with a proper upper bound, while the maximum likelihood terms are already valid terms, as discussed earlier. We note that Theorem~\ref{thm:mlplusplus} is in fact much simpler than its formal enunciation---this is unavoidable, since we are combining different possible bounds for the term $\log\pr(\D_{\vect{n}_{j^\star}}|\vect{\alpha}_{j_S})$ that appears in Lemma~\ref{thm:mlplus} into one bound, while also keeping all the other maximum likelihood bounds. Moreover, to make Theorem~\ref{eq:pp} slightly more compact, we sum all maximum likelihood (ML) terms (first summation in the expression) and then we discard one of them (the first negative ML term) in order to (potentially) replace it with a better bound. This is the only reason why the definition of $\underline{h}$ in the following theorem looks unpleasant to the eyes.

\begin{theorem}[ub$_h$]
Let $S\subseteq\Vi$, $\alpha=\ESS/q(S)$, $j_S\in\getv{\U}{S}$, and 
$\overline{\lbda}_{\vect{n}_{j}}(\alpha) = \lbda_{\vect{n}_{j}}(\alpha)$
if $\alpha\leq 1$ and $\pdv{\lbda_{\vect{n}_j}}{\alpha} (\alpha)\geq 0$, and zero otherwise. Let
\begin{equation*}\label{eq:pp}
    \begin{split}
        & \blbd{S}{j_S} 
         = \sum_{\substack{j\in\getv{\U}{\Vi}:\\ \getv{j}{S}=j_S}} \text{ML}(\vect{n}_{j}) + \min_{\substack{j\in \getv{\U}{\Vi}:\\ \getv{j}{S}=j_S}}\Big(
        -\text{ML}(\vect{n}_{j})\\
        & +\min\{\text{ML}(\vect{n}_{j});~f(\Vi,j)+g(\Vi,j,\alpha);~  \overline{\lbda}_{\vect{n}_{j}}(\alpha)\}\Big).
    \end{split}
\end{equation*}
Then $\lbd{S}{j_S}\leq \blbd{S}{j_S}.$ Moreover,
if $\bd{S'}\geq \sum_{j_S\in \getv{\U}{S}} \blbd{S}{j_S}=\blbdh(S)$ for some $S'\subset S$, then $S$ and all its supersets are not in $\LI_i$.
\label{thm:mlplusplus}
\end{theorem}
\begin{proof}
For parent set $S$, the bound based on $\text{ML}(\vect{n}_{j})$ only (first option in the inner minimisation, which cancels out the double ML terms) is valid by Corollary~\ref{cor:ll}. The other two options rely on Lemma~\ref{thm:mlplus} and their own results: the bound on $f(\Vi,j)+g(\Vi,j,\alpha)$ is valid by Lemma~\ref{theo:cassioprop}, while the bound based on $\overline{\lbda}_{\vect{n}_{j}}(\alpha)$ comes from Lemma~\ref{thm:useconcave}, and thus the result holds for $S$. Take $T\supset S$. It is straightforward that
\begin{equation*}
    \begin{split}
        &\bd{T}\leq \sum_{j_T\in \getv{\U}{T}} \blbd{T}{j_T} =
        \sum_{j_S\in \getv{\U}{S}}\left( \sum_{j_T\in\getv{\U}{T}:~ \getv{j_T}{S}=j_S} \blbd{T}{j_T}\right)
        \leq
        \sum_{j_S\in \getv{\U}{S}} \blbd{S}{j_S},
    \end{split}
\end{equation*}
since $\sum_{j_T\in\getv{\U}{T}:~ \getv{j_T}{S}=j_S} \blbd{T}{j_T}\leq \blbd{S}{j_S}$, because both sides run over the same full instantiations and the right-hand side use the tighter minimisation of Expression~\eqref{eq:pp} only once, while the left-hand side can use that tighter minimisation once every $j_T$, and Lemmas~\ref{theo:cassioprop} and~\ref{thm:useconcave} ensure that the computed values $f(\Vi,j)+g(\Vi,j,\alpha)$ and $\overline{\lbda}_{\vect{n}_{j}}(\alpha)$ are valid for $T$.
\end{proof}

Like previous theorems, Theorem~\ref{thm:mlplusplus} gives us a new upper bound on the local score of a parent set $S$
\begin{equation*}
    \text{ub}_h(S) = \blbdh(S)=\sum_{j_S\in \getv{\U}{S}} \blbd{S}{j_S} \geq \max_{T:T \supset S} \bd{T}.
\end{equation*}

\section{Combining the Bounds}\label{sect:comb}
We note that bound ub$_g$ of the previous section was obtained in a similar way as ub$_f$, and we prove that $\text{ub}_g(S) \leq \text{ub}_f(S)$ for any candidate parent set $S$. Conversely, ub$_h$ bears no such relation to ub$_f$ as we derived it through a new route, studying the properties of the likelihood function.
This is to our advantage, as due to their independent theoretical derivations, ub$_g$ and ub$_h$ prune different regions of the search space and can be effectively combined into a tighter bound $\text{ub}_{g,h}=\min\{\text{ub}_g;\,\text{ub}_h\}$. 

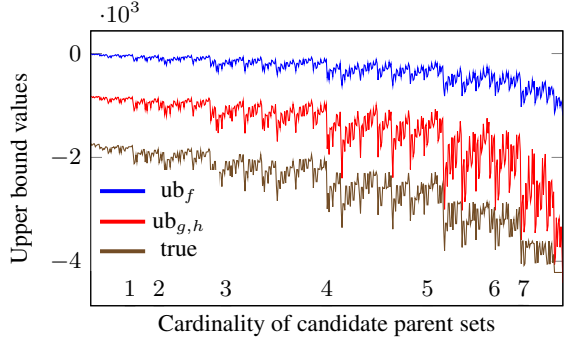
\begin{figure}[ht]
    \centering
    \begin{tikzpicture}[every node/.style={transform shape, font=\footnotesize}
    ]
        \begin{axis}[
            width=0.5625\textwidth,
            height=0.375\textwidth,
            name=first plot,
            every axis plot post/.append style=
			{mark=none},
			y tick label style={
                /pgf/number format/.cd,
                    precision=1,
                /tikz/.cd
            },
            scaled y ticks=base 10:-3,
            enlarge x limits=false,
            cycle list name=color,
            legend style={at={(0.275,0.5)},
                line width=1.5pt,
                draw=none},
            xtick=\empty,
            xlabel=Cardinality of candidate parent sets,
            ylabel=Upper bound values,
            x label style={at={(axis description cs:0.5,0)},anchor=north}
        ]
            \addplot table [x=ind, y=f, col sep=comma] {ub_plot.csv};
            \addplot table [x=ind,y={min}, col sep=comma] {ub_plot.csv};
            \addplot table [x=ind, y={true}, col sep=comma] {ub_plot.csv};
            \legend{ub$_f$, ub$_{g,h}$, true};
        \end{axis}
        \begin{axis}[
            width=0.5625\textwidth,
            height=0.375\textwidth,,
            ymin=0, ymax=10000,
            xtick=\empty,
            ytick=\empty]
            \addplot+[style={color=black, mark=-}, nodes near coords] coordinates {(9, 1) (45, 2) (129, 3) (255, 4) (381, 5) (465, 6) (501, 7)};
        \end{axis}
    \end{tikzpicture}
    \vspace{-0.25cm}
    \caption{Upper bound values for each candidate parent set for variable Standard-of-living-index in the \textit{CMC} dataset \cite{Dua:2019}. Parent sets are arbitrarily ordered within each cardinality (neighbourhood in the graph within same cardinality is not relevant).}  
    \label{fig:ub_values}
    \vspace{-0.4cm}
\end{figure}

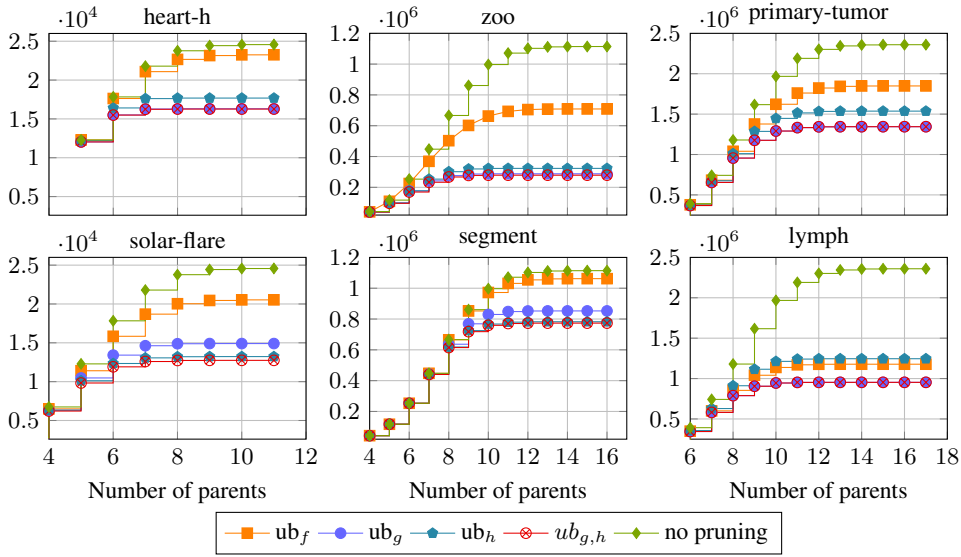
\begin{figure}[h!]
    \centering
    \begin{tikzpicture}
        \begin{groupplot}[
            group style={
                group size=3 by 2,
                x descriptions at=edge bottom,
                vertical sep=16pt,
                horizontal sep=24pt,
            },
            y tick label style={
                /pgf/number format/.cd,
                    fixed,
                    precision=1,
                /tikz/.cd,
                font=\footnotesize 
            },
            grid=both,
            cycle list name=exotic,
            max space between ticks=1000pt,
            try min ticks=5,
            xtick distance=2,
            xlabel = Number of parents,
            footnotesize,
            legend style={at={(1.75,-0.4)},
                anchor=north,legend columns=-1},
        ]
            \nextgroupplot[height=4cm, title=heart-h, xmin=4, xmax=12, ymin=2600, ymax=26000]
                \addplot+[const plot mark left] table [x=palim, y=f, col sep=comma] {Results/complete_heart-h.csv};
                \addplot+[const plot mark left] table [x=palim, y=g, col sep=comma] {Results/complete_heart-h.csv};
                \addplot+[const plot mark left] table [x=palim, y={h}, col sep=comma] {Results/complete_heart-h.csv};
                \addplot+[const plot mark left] table [x=palim, y={min}, col sep=comma] {Results/complete_heart-h.csv};
                \addplot+[const plot mark left] table [x=palim, y={all scores}, col sep=comma] {Results/complete_heart-h.csv}; 
            \nextgroupplot[height=4cm, title=zoo, xmin=4, xmax=17, ymin=20000, ymax=1200000]
                \addplot+ table [x=palim, y=f, col sep=comma] {Results/complete_zoo.csv};
                \addplot+[const plot mark left] table [x=palim, y=g, col sep=comma] {Results/complete_zoo.csv};
                \addplot+[const plot mark left] table [x=palim, y={h}, col sep=comma] {Results/complete_zoo.csv};
                \addplot+[const plot mark left] table [x=palim, y={min}, col sep=comma] {Results/complete_zoo.csv};
                \addplot+[const plot mark left] table [x=palim, y={all scores}, col sep=comma] {Results/complete_zoo.csv};
            \nextgroupplot[height=4cm, title=primary-tumor, xmin=6, xmax=18, ymin=250000, ymax=2500000]
                \addplot+[const plot mark left] table [x=palim, y=f, col sep=comma] {Results/complete_primary-tumor.csv};
                \addplot+[const plot mark left] table [x=palim, y=g, col sep=comma] {Results/complete_primary-tumor.csv};
                \addplot+[const plot mark left] table [x=palim, y={h}, col sep=comma] {Results/complete_primary-tumor.csv};
                \addplot+[const plot mark left] table [x=palim, y={min}, col sep=comma] {Results/complete_primary-tumor.csv};
                \addplot+[const plot mark left] table [x=palim, y={all scores}, col sep=comma] {Results/complete_primary-tumor.csv}; 
            \nextgroupplot[height=4cm, title=solar-flare, xmin=4,xmax=12, ymin=2600, ymax=26000]
                \addplot+[const plot mark left] table [x=palim, y=f, col sep=comma] {Results/complete_solar-flare_2.csv};
                \addplot+[const plot mark left] table [x=palim, y=g, col sep=comma] {Results/complete_solar-flare_2.csv};
                \addplot+[const plot mark left] table [x=palim, y={h}, col sep=comma] {Results/complete_solar-flare_2.csv};
                \addplot+[const plot mark left] table [x=palim, y={min}, col sep=comma] {Results/complete_solar-flare_2.csv}; 
                \addplot+[const plot mark left] table [x=palim, y={all scores}, col sep=comma] {Results/complete_solar-flare_2.csv};
                \legend{{ub$_f$}, {ub$_g$}, {ub$_h$}, {$ub_{g,h}$}, {no pruning}};
            \nextgroupplot[height=4cm, title=segment, xmin=4, xmax=17, ymin=20000, ymax=1200000]
                \addplot+[const plot mark left] table [x=palim, y=f, col sep=comma] {Results/complete_segment.csv};
                \addplot+[const plot mark left] table [x=palim, y=g, col sep=comma] {Results/complete_segment.csv};
                \addplot+[const plot mark left] table [x=palim, y={h}, col sep=comma] {Results/complete_segment.csv};
                \addplot+[const plot mark left] table [x=palim, y={min}, col sep=comma] {Results/complete_segment.csv};
                \addplot+[const plot mark left] table [x=palim, y={all scores}, col sep=comma] {Results/complete_segment.csv};
            \nextgroupplot[height=4cm, title=lymph, xmin=6, xmax=18, ymin=250000, ymax=2500000]
                \addplot+[const plot mark left] table [x=palim, y=f, col sep=comma] {Results/complete_lymph.csv};
                \addplot+[const plot mark left] table [x=palim, y=g, col sep=comma] {Results/complete_lymph.csv};
                \addplot+[const plot mark left] table [x=palim, y={h}, col sep=comma] {Results/complete_lymph.csv};
                \addplot+[const plot mark left] table [x=palim, y={min}, col sep=comma] {Results/complete_lymph.csv};
                \addplot+[const plot mark left] table [x=palim, y={all scores}, col sep=comma] {Results/complete_lymph.csv}; 
       \end{groupplot}
    \end{tikzpicture}
    \caption{Number of scores computed per maximum number of parents with different pruning bounds for four UCI datasets~\cite{Dua:2019} with 12 (heart-h, solar-flare), 17 (zoo, segment) and 18 (primary-tumor, lymph) variables. The scores were computed using breadth-first search.}  \label{fig:n_scores}
\end{figure}

This work focus on new theoretical derivations leading to tighter bounds, and thus an empirical analysis is beyond its scope. Nonetheless, we illustrate possible gains as well as a comparison of the different bounds in simple benchmark datasets in Figures~\ref{fig:ub_values} and \ref{fig:n_scores}. The code for computing these bounds and reproducing the experiments is available on the authors' pages.

For small datasets, it is feasible to score every candidate parent set so that we can compare how far the upper bounds for a given parent set $S$ (and all its supersets) are from the true best score among itself and its supersets. Figure~\ref{fig:ub_values} shows such a comparison for variable \emph{Standard-of-living-index} in the \emph{CMC} dataset \cite{Dua:2019}, which has 10 variables and 1,473 instances. It is clear that the new bound $\text{ub}_{g,h}$ is much tighter than the current best bound in the literature (here called ub$_f$) and improves considerably towards the true best score (only known because this particular dataset is not too large).

For larger datasets (more than 10 variables), evaluating all candidate parent sets becomes computationally impracticable, so instead we evaluate the number of scores computed with each bound. In Figure~\ref{fig:n_scores}, we see the new bounds considerably reduced the number of scores computed, which translates into smaller lists of potentially optimal parent sets $L_i$ (see Definition~\ref{def:psi}). This goes to show the practical value of tighter upper bounds, as we save computing time in both steps of BNSL: parent set identification (fewer scores to compute) and structure optimisation (smaller search space).

Finally, we point out that the mathematical results may seem harder to apply than they actually are. Computing ub$_g(S)$ and ub$_h(S)$ to prune a parent set $S$ and all its supersets can be done
in linear time, as one pass through the data is enough to collect and process
all required counts; more sophisticated data structures, such as AD-trees \cite{moore98:_cached_suffic_statis_effic_machin}, might allow for even greater speedups. Since calculating a score already takes linear time in the number of data samples, we have cheap bounds which are provably superior to the current state-of-the-art pruning for BDeu.

\section{Conclusions}\label{sect:conc}
We introduced new theoretical upper bounds for exact structure learning of Bayesian networks with the BDeu score by studying the score function from multiple angles. These bounds are provably tighter than previous results and shall provide significant benefits in reducing the search space in candidate parent set identification in BNSL and potentially other applications involving independence assumptions.

A natural step for future research is the integration of our bounds with more sophisticated data structures and search algorithms. As an example, branch-and-bound methods are particularly promising as they not only consider the parent sets and its corresponding full instantiations but also partial instantiations that are formed by disallowing some variables to be parents in some of the branches. Our results also open new routes for further theoretical work in exact structure learning. Notably, we conjecture that the maximum likelihood estimation terms still leave room for tighter bounds.

\bibliographystyle{plain}
\bibliography{jc}

\newpage
\section*{Appendix - Further Experimental Results}

\pgfplotsforeachungrouped \i in {Results/final_diabetes.csv,
                                Results/final_nursery.csv,
                                Results/final_cmc.csv,
                                Results/final_heart-h.csv,
                                Results/final_solar-flare_2.csv,
                                Results/final_vowel.csv,
                                Results/final_zoo.csv,
                                Results/final_vote.csv,
                                Results/final_segment.csv,
                                Results/final_pendigits.csv,
                                Results/final_lymph.csv,
                                Results/final_primary-tumor.csv,
                                Results/final_vehicle.csv,
                                Results/final_hepatitis.csv,
                                Results/final_colic.csv,
                                Results/final_autos.csv,
                                Results/final_flags.csv} 
                                {%
    \pgfplotstablevertcat{\output}{\i}
}
\setlength{\tabcolsep}{4pt}
\begin{table*}[h!]
    \centering
    \fontsize{8}{8}\selectfont
    \caption{Number of computations pruned ($|L^c|=|\text{search space}|-|L|$) with each bound: ub$_f$, ub$_g$, ub$_h$ and $ub_{g,h}$. Each dataset is characterised by its number of variables and observations, $n$ and $N$, and the number of all possible parent combinations $|\text{search space}|$. The maximum imposed in-degree is given by in-d and $\frac{|\text{ub}_g<\text{ub}_h|}{|\text{ub}_h<\text{ub}_g|}$ is the proportion of times ub$_g$ was tighter than ub$_h$.}
    \pgfplotstabletypeset[
        precision=3,
        every head row/.style={before row=\toprule, after row=\midrule},
        every last row/.style={after row=\bottomrule},
        every nth row={3}{before row=\specialrule{0.001em}{0.08em}{0.08em}},
        columns/palim/.style={column name=in-d,
            assign cell content/.code={
            \pgfmathparse{int((\pgfplotstablerow)}
            \ifnum\pgfmathresult=2\pgfkeyssetvalue{/pgfplots/table/@cell content}{$\infty$}
            \fi
            },
        },
        columns/|S|/.style={column name=$|\text{search space}|$},
        columns/Dataset/.style={string type, column type=p{1.05cm}|,
            assign cell content/.code={\pgfkeyssetvalue{/pgfplots/table/@cell content}{\multirow{3}{1.05cm}{##1}}%
            },
        },
        columns/naive/.style={column name=$|L^c_f|$},
        columns/naive++/.style={column name=$|L^c_{g}|$},
        columns/{new bound}/.style={column name=$|L^c_{h}|$},
        columns/{min}/.style={column name={$|L^c_{g,h}|$}},
        columns/{g/h}/.style={column name={$\frac{|\text{ub}_g<\text{ub}_h|}{|\text{ub}_h<\text{ub}_g|}$}},
        columns={Dataset, n, N, |S|, palim, naive, naive++, {new bound}, min, g/h}
    ]{\output}
    \label{tab:results}
\end{table*}

\end{document}